\documentclass[sigconf]{acmart}
\AtBeginDocument{%
  }

\usepackage{mathrsfs}
\usepackage{mathtools}
\usepackage{url}            
\usepackage{bm}
\usepackage{nicefrac}       
\usepackage{adjustbox}
\usepackage{blindtext}
\usepackage{enumitem}
\usepackage{listings}
\usepackage{pifont}
\usepackage{multirow}
\usepackage{multicol}
\usepackage{makecell}
\usepackage{algorithm}
\usepackage{rotating}
\usepackage{algorithm}
\usepackage{algorithmic}
\usepackage{colortbl}
\usepackage{soul}
\usepackage{tcolorbox}
\usepackage[capitalize,noabbrev]{cleveref}
\definecolor{babyblue}{rgb}{0.54, 0.81, 0.94}
\definecolor{bisque}{rgb}{1.0, 0.89, 0.77}
\definecolor{bshade}{rgb}{0.55,0.75,0.95}

\definecolor{mygray}{gray}{.6}
\definecolor{myblue}{RGB}{89,158,254}
\definecolor{mygreen1}{RGB}{81,150,111}
\definecolor{mygreen2}{RGB}{93,174,86}
\definecolor{myred}{RGB}{160,0,0}
\definecolor{myyellow}{RGB}{227,207,87}

\definecolor{mygray}{gray}{.6}
\definecolor{myblue}{RGB}{89,158,254}
\definecolor{mygreen1}{RGB}{81,150,111}
\definecolor{mygreen2}{RGB}{93,174,86}
\definecolor{myred}{RGB}{160,0,0}
\definecolor{myyellow}{RGB}{227,207,87}
\usepackage{caption}
\usepackage[capitalize,noabbrev]{cleveref}
\usepackage{threeparttable}
\usepackage{wrapfig}

\theoremstyle{plain}

\theoremstyle{definition}

\theoremstyle{remark}

\definecolor{babyblue}{rgb}{0.54, 0.81, 0.94}
\definecolor{bisque}{rgb}{1.0, 0.89, 0.77}
\definecolor{bshade}{rgb}{0.55,0.75,0.95}

\definecolor{mygray}{gray}{.6}
\definecolor{myblue}{RGB}{89,158,254}
\definecolor{mygreen1}{RGB}{81,150,111}
\definecolor{mygreen2}{RGB}{93,174,86}
\definecolor{myred}{RGB}{160,0,0}
\definecolor{myyellow}{RGB}{227,207,87}
\usepackage{bbding}

\usepackage{enumitem}
\let\oldding\ding
\renewcommand{\ding}[2][1]{\scalebox{#1}{\oldding{#2}}}
\usepackage[textsize=tiny]{todonotes}

\newcommand{\cc}{\color[rgb]{0,0.6,0.3}\checkmark}
\newcommand{\xx}{\color[rgb]{0.6,0,0}{\ding{55}}}

\copyrightyear{2025}
\acmYear{2025}
\setcopyright{acmlicensed}\acmConference[MM '25]{Proceedings of the 33rd
ACM International Conference on Multimedia}{October 27--31, 2025}{Dublin, Ireland}
\acmBooktitle{Proceedings of the 33rd ACM International Conference on
Multimedia (MM '25), October 27--31, 2025, Dublin, Ireland}
\acmDOI{10.1145/3746027.3754781}
\acmISBN{979-8-4007-2035-2/2025/10}

\begin{document}

\title{FocusTrack: One-Stage Focus-and-Suppress Framework for 3D Point Cloud Object Tracking}





\author{Sifan Zhou}
\affiliation{
  \institution{School of Automation\\ Southeast University,}
  \institution{Key Laboratory of Measurement and Control of Complex Systems of Engineering, Ministry of Education}
  \city{Nanjing}
  \country{China}
}
\email{sifanjay@gmail.com}

\author{Jiahao Nie}
\affiliation{
  \institution{School of Information Technology and Artificial Intelligence, Zhejiang University of Finance and Economics}
  \city{Hangzhou}
  \country{China}
}
\email{jhnie@zufe.edu.cn}

\author{Ziyu Zhao}
\affiliation{
  \institution{School of Automation\\ Southeast University,}
  \institution{Key Laboratory of Measurement and Control of Complex Systems of Engineering, Ministry of Education}
  \city{Nanjing}
  \country{China}
}
\email{zhaoziyu.950207@gmail.com}

\author{Yichao Cao}
\affiliation{
  \institution{Big Data Institute\\Central South University}
  \city{Changsha}
  \country{China}
}
\email{caoyichao@csu.edu.cn}

\author{Xiaobo Lu}
\authornote{Corresponding Author.}
\affiliation{
  \institution{School of Automation\\ Southeast University,}
  \institution{Key Laboratory of Measurement and Control of Complex Systems of Engineering, Ministry of Education}
  \city{Nanjing}
  \country{China}
}
\email{xblu2013@126.com}


\renewcommand{\shortauthors}{Sifan Zhou, Jiahao Nie, Ziyu Zhao, Yichao Cao, \& Xiaobo Lu}
\begin{abstract}
In 3D point cloud object tracking, the motion-centric methods have emerged as a promising avenue due to its superior performance in modeling inter-frame motion. However, existing two-stage motion-based approaches suffer from fundamental limitations: (1) \textbf{error accumulation} due to decoupled optimization caused by explicit foreground segmentation prior to motion estimation, and (2) \textbf{computational bottlenecks} from sequential processing. To address these challenges, we propose \textbf{FocusTrack}, a novel one-stage paradigms tracking framework that unifies motion-semantics co-modeling through two core innovations: \textbf{Inter-frame Motion Modeling (IMM)} and \textbf{Focus-and-Suppress Attention}. The IMM module employs a temp-oral-difference siamese encoder to capture global motion patterns between adjacent frames. The Focus-and-Suppress attention that enhance the foreground semantics via motion-salient feature gating and suppress the background noise based on the temporal-aware motion context from IMM without explicit segmentation. Based on above two designs, FocusTrack enables end-to-end training with compact one-stage pipeline. Extensive experiments on prominent 3D tracking benchmarks, such as KITTI, nuScenes, and Waymo, demonstrate that the FocusTrack achieves new SOTA performance while running at a high speed with 105 FPS.
\end{abstract}





\maketitle
\vspace{-2mm}
\section{Introduction}
\label{sec:intro}

Environments perception is important for autonomous driving and mobile robots\cite{zhao2026advances,zhao2024balf,lu2025yo,yan2025turboreg,ma2025reinforcement,liu2025pipe,li2024mlp,hou2025fedc,ma2025energy, lu2023tf, wang2024scantd, wang2025target,lu2024mace,sun2024gsrender,zhao2025tartan,shi2025rethinking}. Among them, Single object tracking (SOT) plays a key role in various computer vision applications, such as autonomous driving, visual surveillance and robotics~\cite{javed2022visual,zhang2020empowering,cui2019point}. Early efforts have predominantly focused on RGB images obtained by cameras~\cite{siamfc,ocean,mixformer}. 
In recent years, with the development of LiDAR sensors, many techniques~\cite{p2b,glt,v2b,cxtrack,m2track} for 3D SOT have been presented.

\begin{figure}[h]
\centering
\includegraphics[width=1.0\linewidth]{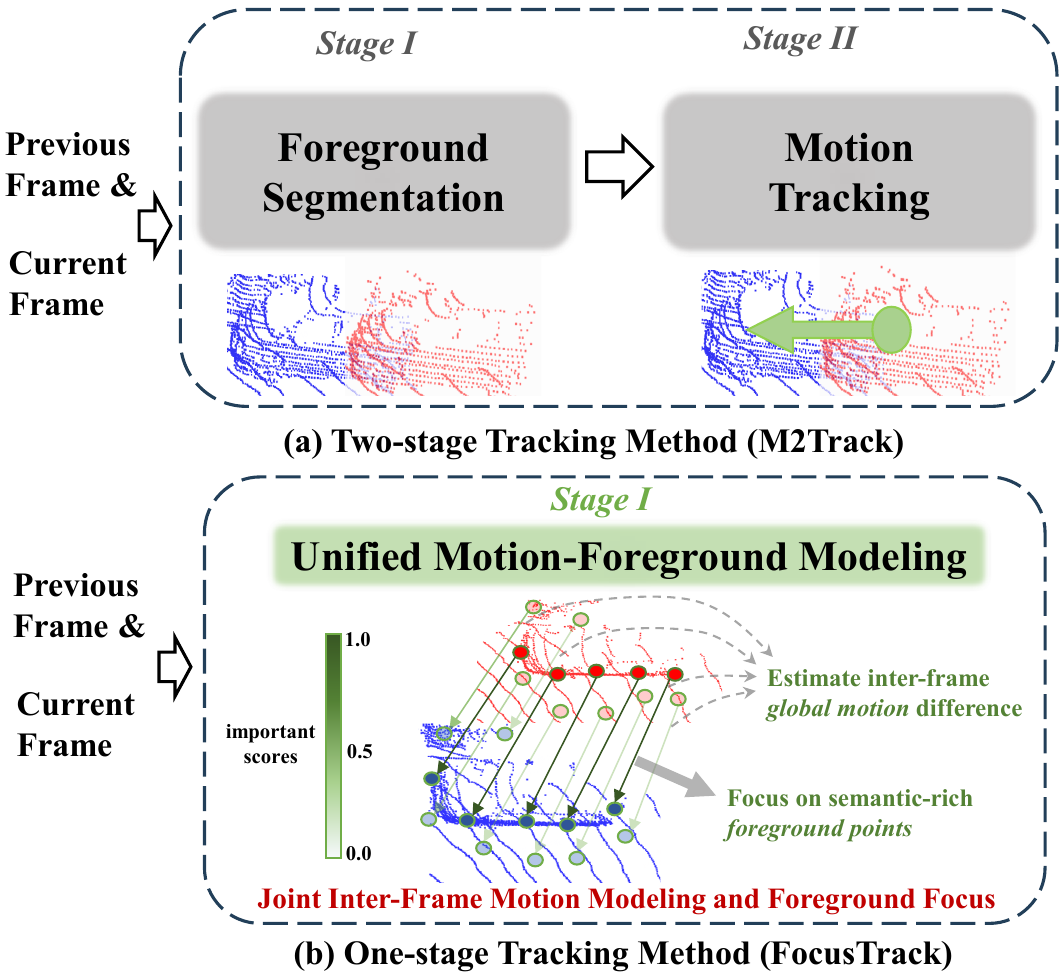}
\vspace{-3mm}
\caption{Comparison between two-stage motion-based tracking methods (a) and our two-stage motion-based tracking method (b). The two-stage motion-based methods M2Track~\cite{m2track,m2track++} series models motion relation for tracking in a two-stage manner. In contrast, our FocusTrack explores unified motion-foreground modeling for one-stage tracking.}
\vspace{-6mm}
\label{fig:general}
\end{figure}

Current prevailing 3D SOT methods~\cite{sc3d,p2b,bat,cui2019point,3dsiamrpn,stnet,pillartrack} primarily follow an appearance matching-based framework. This framework consists of two key modules:  1) a Siamese-like backbone to extract features of template and search region and 2) a feature interaction module to enhance correlation. Despite the advanced performance of this matching-based paradigm, it often neglects the local motion information of the target across frames. Additionally, point clouds inherent sparsity and occlusion present significant challenges for appearance matching.   %

Recently, motion-centric tracking framework has emerged as a promising solution in 3D SOT~\cite{m2track, m2track++}, demonstrating superior accuracy. This framework reformulates the object tracking problem as relative motion regression across successive frames. Benefiting from it, the model can reduce its severe reliance on the appearance of point clouds, enabling more robust tracking results in sparse or occluded environments. However, existing motion-centric methods~\cite{m2track, p2p, li2024flowtrack} generally treat all inter-frame motion differences uniformly, overlooking the fact that points in different spatial locations exhibit distinct motion patterns. Foreground targets typically exhibit more pronounced dynamics changes and inherent motion patterns, whereas background points tend to remain relatively static. This discrepancy highlights a critical gap in current motion-centric approaches: they inadequately account for the varying significance of motion differences across frames. Furthermore, background noise can interfere with the modeling of foreground motion, complicating the tracking process. Even for M2Track~\cite{m2track,m2track++}, which employs a "segmentation-and-motion prediction" pipeline, a refinement module is still needed to compensate the errors caused by background noise. Besides, their separated two-stage manner with decoupled optimization further limited the performance and efficiency.

\begin{figure}[t]
\centering
\includegraphics[width=1.0\linewidth]{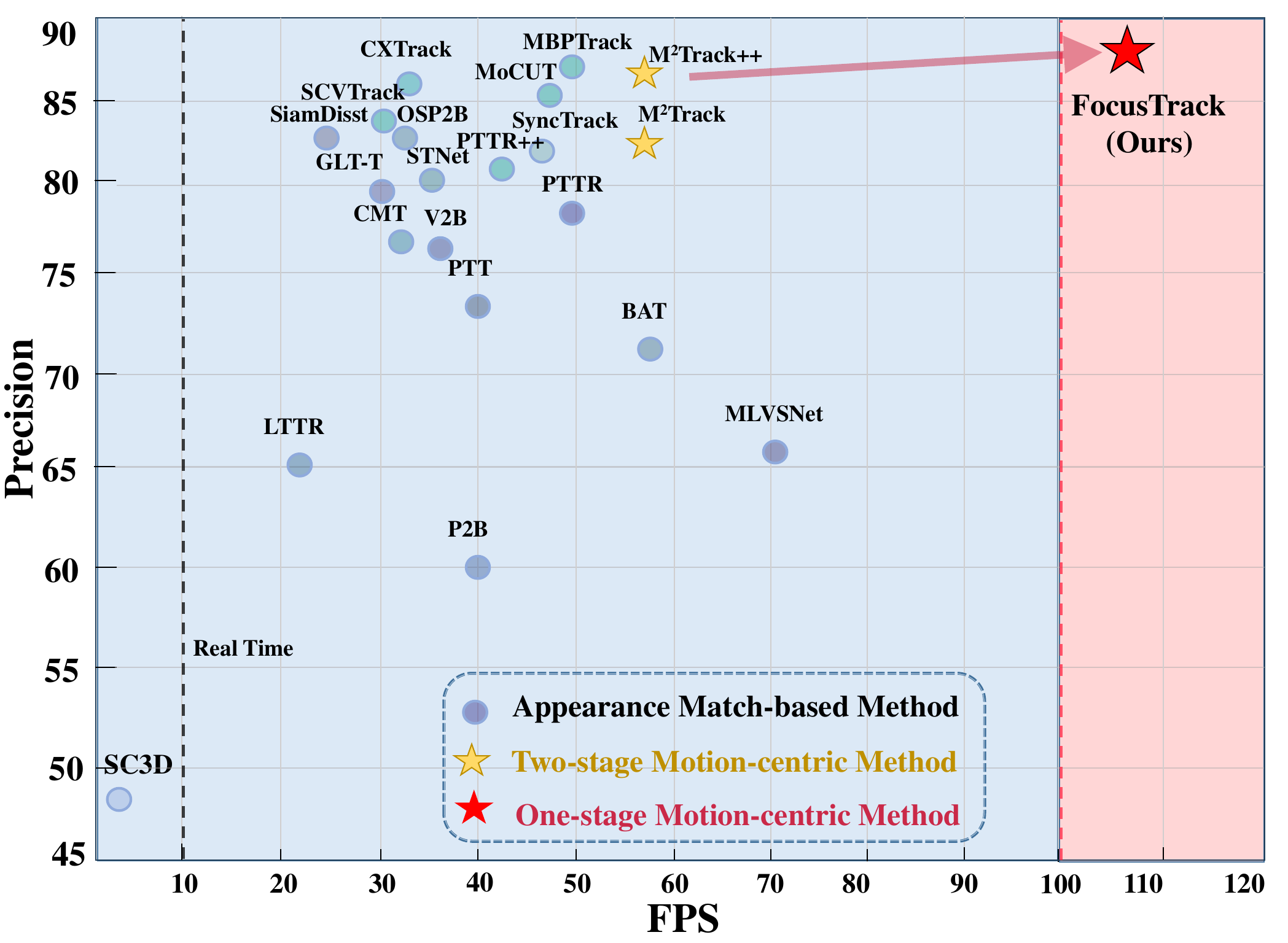}
\vspace{-5mm}
\caption{Comparison with state-of-the-art methods. We visualize mean precision across all categories on KITTI dataset~\cite{kitti} with respect to running speed (FPS). FocusTrack shows the new SOTA in terms of both accuracy and speed.}
\vspace{-6mm}
\label{fig1}
\end{figure}

Above limitations stem from a core oversight: Motion modeling should discriminate semantically meaningful movements from static noise, rather than treating all points equally. As shown in Fig.~\ref{fig:general}, we propose \textbf{FocusTrack}, a novel one-stage motion-based tracking framework that jointly optimize motion and geometry semantics based on the point clouds from the previous and current frames. Specifically, FocusTrack is composed of FocusTrack blocks, which incorporate two key designs:(1) \textbf{Inter-frame Motion Modeling (IMM)} module and (2) \textbf{Focus-and-Suppress} attention. In IMM module, we employs a temporal-difference siamese network to capture the inter-frmae global motion differences. The \textbf{Focus-and-Suppress Attention} aims to extract the intra-frame spatial geometric information based on the temporal-aware
motion context from IMM without explicit segmentation. In detail, the output of IMM is formulated as a weighted map, providing a temporal-aware motion context for the attention mechanism, enabling it to weight the spatial semantics for enhancing the foreground and suppressing the background noise. Extensive experiments on KITTI~\cite{kitti}, nuScenes~\cite{nuScenes} and WOD~\cite{waymo} datasets demonstrate that FocusTrack sets a new state-of-the-art (SOTA) performance, achieving a high speed of 105 FPS on a single RTX3090 GPU. FocusTrack significantly outperforms the previous top-performing match-based method~\cite{mbptrack} across all datasets and surpasses the current SOTA motion-based method~\cite{p2p} by $\sim2\%$ on large-scale nuScenes and WOD datasets.

Notably, while previous studies~\cite{ptt,ptt-journal,gltt} have also explored the transformer for 3D SOT, they typically are stuck in the performance-limited matching paradigms. In contrast, our FocusTrack aims to effectively leverage inter-frame differential motion information within a motion-centric framework to enhance the semantics of foreground objects. As shown in Fig.~\ref{fig1}, our approach demonstrates greater efficiency while maintaining SOTA performance. The main contributions are summarized as follows:
\begin{enumerate}
    \item \textbf{Inter-frame Motion Modeling (IMM):} IMM models the inter-frame global motion differences, enabling the effective capture and utilization of inter-frame motion.
    \item \textbf{Focus-and-Suppress Attention:} Focus-and-Suppress Attention focuses on enhancing the foreground semantics based on the temporal-aware motion context.
    \item \textbf{One-Stage Co-optimization Framework:} The FocusTrack unifies motion-semantics co-optimization into an one-stage paradigm, significantly improving performance and computational efficiency.
    \item \textbf{SOTA Performance and Speed:} Extensive experiments on prominent 3D tracking benchmarks, such as KITTI, nuScenes, and Waymo, demonstrate that the FocusTrack achieves new SOTA performance while running at 105 FPS.
\end{enumerate}

\section{Related Work}
\label{sec:related}
\subsection{Siamese-based 3D Single Object Tracking}

Early 3D SOT methods regard object tracking as appearance matching issue. They use the Siamese network for feature extraction and reveal the local tracking cues based on the similarity matching. As a pioneer, SC3D~\cite{sc3d} introduces the first work to integrate the Siamese network to extract features from template and candidates and then compute the feature distances to select suitable tracking result. Subsequently, 3D-SiamRPN~\cite{3dsiamrpn} employs a region proposal network (RPN) into the 3D Siamese paradigm. P2B~\cite{p2b} utilizes VoteNet~\cite{votenet} to generate a batch of high-quality candidates with an end-to-end framework. Inspired by P2B, numerous networks~\cite{fan2026beyond,cmt,osp2b,glt,MVCTrack} have been adopted the same Siamese paradigm. For instance, BAT~\cite{bat} encodes the object's size priors to augment correlation learning between the template and search areas. PTT~\cite{ptt,ptt-journal}, LTTR~\cite{lttr}, CMT~\cite{cmt}, and STNet~\cite{stnet} explore different attention mechanisms to improve feature propagation and correlation. CAT~\cite{CAT}, STTracker~\cite{cui2023sttracker} and SCVTrack~\cite{robust} explore the temporal and completion to improve the accuracy on sparse scenes. Recently, CXTrack~\cite{cxtrack} emphasizes the importance of context for tracking performance by a target-centric transformer. MBPTrack~\cite{mbptrack} designs an external memory to enhance the spatial and temporal information aggregation. Despite the superior performance, the matching-based paradigm often overlooks the local motion information of the target across frames. Furthermore, point clouds are typically incomplete and lack texture, their inherent sparsity and occlusion present challenges to appearance matching.

\begin{figure*}[t]
\centering
\includegraphics[width=0.99\linewidth]{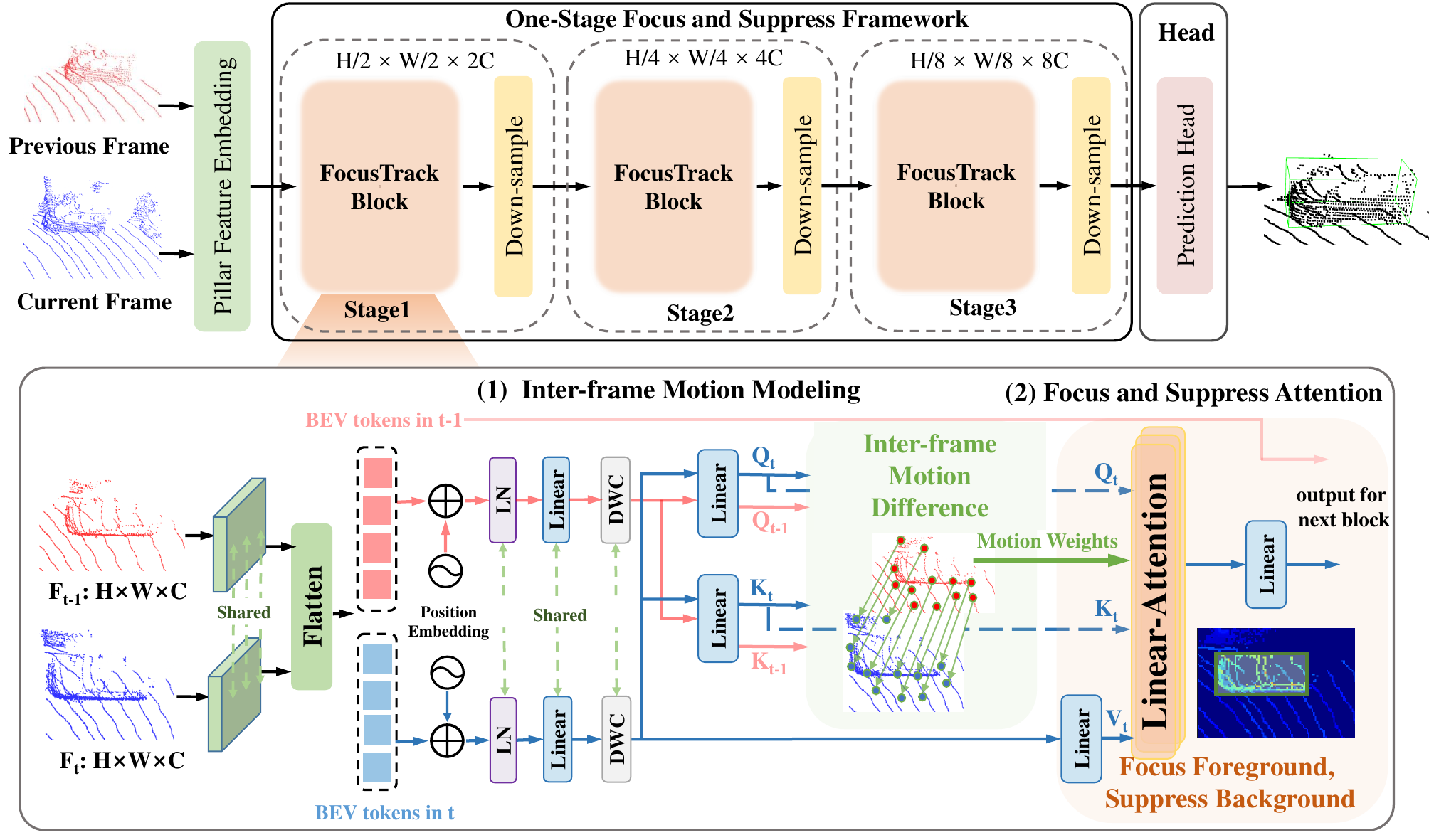}
\vspace{-2mm}
\caption{One-stage focus-and-suppress framework (\textbf{FocusTrack}): This framework consists of a backbone that capture inter-frame differences for motion modeling. Then, it features a Focus-and-Suppress attention mechanism designed to suppress background noise while enhancing foreground semantics based on motion weights.}
\label{fig:framework}
\vspace{-3mm}
\end{figure*}

\vspace{-2.5mm}

\subsection{Motion-based 3D Single Object Tracking}
In addition to appearance matching-based trackers, there are efforts to model inter-frame motion. M$^2$Track~\cite{m2track, m2track++} proposes a motion-centric tracking framework that first segments the foreground points of the target from the search regions, and then infers the target's 4-DOF relative motion offset. Furthermore, M$^2$Track++~\cite{m2track++} further explore the promising potential of the motion-based paradigm in semi-supervised application. Similarly, VoxelTrack~\cite{lu2024voxeltrack} explore the 3D spatial information from multi-level voxel representation. P2P~\cite{p2p} demonstrate the superiority of motion-centric approach by transforming the complicated appearance matching into the estimation of relative offset between adjacent frames. DMT~\cite{dmt} introduces a motion prediction module that estimates the potential center of the target based on historical bbox and further refines it. Unlike the instance-level modeling in DMT, FlowTrack~\cite{li2024flowtrack} proposes point-level flow to capture the fine-grained motion details based on multiple historical frames. Although achieving promising results, these methods tend to uniformly rely on global motion differences when modeling relative motion, neglecting significant noise disturbances introduced by background. Therefore, in this paper, we propose a new one-stage focus-and-suppress framework that mitigates background noise while enhancing the local motion cues of foreground based on the global motion differences.

\section{Methodology}
\label{sec:method}

\subsection{Task Definition}
We define the task of LiDAR-based 3D Single Object Tracking (3D SOT) as follows: given an initial template point cloud $\mathcal{P}^{t}=\{p_i^t\}_{i=1}^{N_t}$ and its 3D bounding box (BBox) $\mathcal{B}_{t} (x_t,y_t,z_t,w_t,h_t,l_t,\theta_t)$, where $(x,y,z)$ and $(w,h,l)$ denotes the center and size, $\theta$ is the rotation angle around $up$-axis. The goal of 3D SOT is to locate the object in the search region $\mathcal{P}^{s}=\{p_i^s\}_{i=1}^{N_s}$ and output a 3D BBox $\mathcal{B}_{s}$ frame by frame, with $N_t$ and $N_s$ as the number of points in the template and search region. Following ~\citep{mbptrack,ptt,bat}, for both rigid or non-rigid objects, since the size of target remains approximately unchanged in 3D SOT across all frames, predicting only 4 parameters $(x_s,y_s,z_s,\theta_s)$ to represent $\mathcal{B}_{s}$. 

\paragraph{Input and Tracking Process} Following~\cite{cxtrack,p2p}, we use the target and its surrounding context as a template to guide localization in the search region. Similar to\cite{m2track,mbptrack}, we crop point cloud from the previous frame $t-1$ and current frame $t$, centered on the previous prediction$\mathcal{B}_{t-1}$, with an extended spatial range [($x_{min}$, $x_{max}$), ($y_{min}$, $y_{max}$), ($z_{min}$, $z_{max}$)] to obtain $\mathcal{P}^{t-1}$ and $\mathcal{P}^{t}$. The FocusTrack then predict the relative motion between frames to determine the current 3D BBox. This process is formulated as:
\vspace{-2mm}
\begin{equation}
\begin{aligned}
    \label{eq1}
    (\Delta x_t, \Delta y_t, \Delta z_t, \Delta \theta_t) = {\text{FocusTrack}}(\mathcal{P}^{t-1}, \mathcal{P}^{t})
    \end{aligned}
\end{equation}
\begin{equation}
\begin{aligned}
    \label{eq2}
     (x_t,y_t,z_t,\theta_t) = (x_{t-1},y_{t-1},z_{t-1},\theta_{t-1}) \\
     + (\Delta x_t, \Delta y_t, \Delta z_t, \Delta \theta_t)
    \end{aligned}
\end{equation}
where FocusTrack, detailed in Fig.\ref{fig:framework}, includes pillar feature embedding, an inter-frame motion-based backbone for enhancing foreground semantics, and a prediction head. 

\subsection{One-Stage Focus-and-Suppress Framework}

\paragraph{\textbf{Pillar Feature Embedding}}
We use the pillar feature encoder from PointPillars~\cite{pointpillars} to transform sparse input points into dense bird's eye view (BEV) features. Notably, other pillar feature encoding methods~\cite{zhou2023fastpillars,pillarhist} can also be utilized. After pillar feature embedding, the input points $\mathcal{P}^{t-1}$ and $\mathcal{P}^{t}$ are converted into dense BEV features $\mathcal{F}_{t-1}$ and $\mathcal{F}_{t}$ with dimensions $H \times W \times C$, encoding the sparse points into a structured grid. This choice is supported by recent research\cite{stnet,mbptrack,p2p}, which show that grid-based methods outperform point-based approaches~\cite{ptt,pttr,synctrack} in accuracy and speed, offering a more compact and efficient representation compared to point-based encoders~\cite{pointnet,pointnet++}. 

The BEV features $\mathcal{F}_{t-1}$ and $\mathcal{F}_{t}$ through multiple FocusTrack blocks, each followed by a convolution layer with a stride 2, halving spatial resolution and doubling channel count. This process is repeated 3 times, resulting in final features of size $\frac{H}{8} \times \frac{W}{8} \times 8C$, which are then input to the motion prediction head. Similar to ~\cite{synctrack,osp2b}, feature extraction and interaction occur simultaneously in our method, enabling direct motion prediction.

\subsubsection{\textbf{Inter-frame Motion Modeling (IMM)}}
In object tracking, the motion of targets generates significant feature changes between adjacent frames. Effectively capturing and leveraging these changes enbales the model to focus on moving foreground objects, enhancing its semantics understanding across diverse scenes. If inter-frame variations are overlooked, the model risks distributing attention uniformly across global differences, encompassing both foreground and background elements. However, backgrounds often introduce excessive noise, obscuring critical targets. By modeling these temporal changes, the model can inferthe velocity or position offset of targets, facilitating predictions of their future states. The similar insights have also been discussed in methods~\cite{m2track,m2track++,p2p}. Therefore, we aim to leverage the motion differences between adjacent frames effectively, integrating them into the feature extraction and interaction processes to develop an efficient tracking framework.

To achieve this, we introduce an \textbf{Inter-frame Motion Modeling (IMM)} module into the attention mechanism to estimate global motion difference between consecutive frames. Our approach is inspired by the frame differencing algorithm~\cite{radzi2014extraction,ellenfeld2021deep,zhao2022survey}, a well-established technique in video processing for detecting moving objects. In 2D video analysis, frame differencing calculates pixel-wise differences between adjacent frames to highlight regions of significant change, effectively isolating dynamic objects from static backgrounds. We hypothesize that this principle can be adapted to 3D point cloud data, where object motion manifests as spatial and feature displacements across frames. However, directly applying frame differencing to point clouds poses challenges due to their sparse and irregular structure, unlike the dense, grid-like nature of video frames. To address this, we innovate by transferring the concept of frame differencing into a high-dimensional feature space within the attention mechanism, enabling the capture of complex inter-frame relationships tailored to point cloud characteristics.

As shown in Fig.\ref{fig:framework}, the BEV features $\mathcal{F}_{t-1}$ and $\mathcal{F}_{t}$, derived from Pillar feature embedding, are processed through a shared CNN and flattened into BEV tokens $\mathcal{X}_{t-1} \in \mathbb{R}^{N \times C}$ and $\mathcal{X}_{t} \in \mathbb{R}^{N \times C}$. This is expressed as:
\vspace{-2mm}
\begin{equation}
\begin{aligned}
(\mathcal{X}_{t-1}, \mathcal{X}_{t}) = f(\rm CNN (\mathcal{F}_{t-1}, \mathcal{F}_{t}))
\end{aligned}
\label{eq4}
\end{equation}
where $\rm CNN$ is a shared convolution layer to enhance the spatial semantics information 
of the BEV feature and $f$ is the flatten operation to convert to sequential BEV tokens for the attention block. Follwoing~\cite{3DETR}, we adopt a Pre-LN design for the BEV tokens with a layer normalization~\cite{layernormalization} operation $\rm LN$ before modeling the global motion difference. Specifically, we also employ a shared linear layer $\ell$ and a shared depth-wise convolution (DWC) to enhance the spatial and local feature interactions while preserving feature diversity. This process is written as:
\begin{equation}
\begin{aligned}
(\bar{\mathcal{X}}_{t-1}, \bar{\mathcal{X}_{t}}) = {\ell({\rm DWC} (\rm LN (\mathcal{X}_{t-1}, \mathcal{X}_{t})))} 
\end{aligned}
\label{eq4}
\end{equation}
Positional embeddings are added before the normalization (omitted for brevity). These tokens are then mapping to queries and keys:
\vspace{-2mm}
\begin{equation} 
\label{eq:general_attn}
    \begin{aligned}
        Q_{t-1}=\bar{\mathcal{X}}_{t-1}W_Q, K_{t-1}=\bar{\mathcal{X}}_{t-1}W_K,\\
        Q_{t}=\bar{\mathcal{X}}_{t}W_Q, K_{t}=\bar{\mathcal{X}}_{t}W_K,\\
    \end{aligned}
\end{equation}
where $ W_{Q/K}\in\mathbb{R}^{C \times d} $ are projection matrices. The core innovation of IMM lies in modeling motion differences using the attention mechanism, inspired by frame differencing. In traditional attention, $QK^T$ computes feature similarity, reflecting alignment between token representations. Drawing from frame differencing, we propose computing the difference $Q_{t}K_{t}^T - \alpha Q_{t-1}K_{t-1}^T$ to capture the evolution of feature alignments across frames. Here, $Q_{t}K_{t}^T$ represents the current frame’s feature relationships, while $Q_{t-1}K_{t-1}^T$ encodes the previous frame’s state. Subtracting these terms highlights changes in feature similarity, mirroring how frame differencing isolates pixel variations to detect motion. This difference intuitively reflects global motion patterns in the feature space, emphasizing regions with significant shifts—typically moving objects. The learnable parameter $\alpha$ adaptively scales the previous frame’s contribution, accommodating varying motion dynamics (e.g., slow vs. fast-moving objects). To enhance the model’s capacity to learn complex motion patterns, we apply ${\rm SiLu}$ activation function, introducing non-linearity. The resulting motion-aware weights $W_{m}$ are formulated as:

\vspace{-2mm}
\begin{equation} 
\label{eq:motion_sub}
    \begin{aligned}
    W_{m} = {\rm SiLU} (Q_{t}K_{t}^T - \alpha Q_{t-1}K_{t-1}^T),  
    \end{aligned}
\end{equation}
where $\alpha$ is trainable, ${\rm SiLu}$ ensures robust feature modulation. The weights $W_{m}$ could regard as a motion-aware weights, which dynamically adjust the attention output, amplifying focus on dynamic objects while suppressing static background noise. This design is intuitive and effective for point cloud tracking. In point clouds, an object’s motion between frames manifests as coordinate offsets, which, after high-dimensional mapping, translate into feature space shifts. Subtracting previous frame features from current ones captures these global motion differences, akin to frame differencing’s pixel subtraction. By embedding this within the attention mechanism, IMM leverages the model’s ability to learn contextual relationships, overcoming the sparsity and irregularity of point clouds that render direct frame differencing impractical. This approach not only enhances performance by prioritizing dynamic targets but also introduces a novel bridge between classical video processing techniques and modern 3D perception, underscoring the originality of our method.

\subsubsection{\textbf{Hierarchical Focus-and-Suppress Attention}}

By introducing inter-frame motion differences in IMM module, we are able to capture the motion characteristics of target objects across consecutive frames, thereby to better represent rich semantics of dynamic targets. This approach not only aligns with the natural tendency of human visual systems to focus on moving objects, but it also improves the model's attention to foreground dynamic entities. 

Next, we formulate above process. With an input of $N$ BEV tokens represented as $\mathcal{X}\in\mathbb{R}^{N \times C}$, a generalized attention can be formulated as follows in each head:
\begin{equation} 
\label{eq:general_attn}
    \begin{aligned}
        Q=\mathcal{X}W_Q, K=\mathcal{X}W_K, V=\mathcal{X}W_V, \\ 
        O_i=\sum_{j=1}^{N}\ \frac{{\rm Sim}{\left(Q_i,K_j\right)}}{\sum_{j=1}^{N}\ {\rm Sim}{\left(Q_i,K_j\right)}}V_j,
    \end{aligned}
\end{equation}
where $ {\rm Sim}{\left(\cdot,\cdot \right)}$ denotes the similarity function. When ${\rm Sim}\left(Q,K\right)={\rm exp}({QK^T}/{\sqrt d})$ in Eq.\ref{eq:general_attn}, it becomes self-attention~\cite{attention}, which has been highly successful in much previous 3D SOT methods~\cite{mbptrack,cxtrack,synctrack}. However, self-attention requires computing the similarity between all query-key pairs, means $(QK^T)$ leading to $\mathcal{O}(N^2)$ complexity with $ Q/K \in\mathbb{R}^{N \times d}$. Comparatively, linear attention~\cite{linear_attn} efficiently addresses the computation challenge with a linear complexity of $\mathcal{O}(N)$. Then in our FocusTrack, the motion differences weighted linear attention as follows:
\vspace{-2mm}
\begin{equation} 
    \begin{aligned}
        \mathcal{F}_t^{FocusT} & = \ell({\rm FocusT}(\mathcal{F}_{t-1}, \mathcal{F}_t)) \\
        & =\ell({\rm SiLU}(Q)({\rm SiLU}(K_t)^TV_t) \cdot \sigma(\ell(W_m)))\\
    \end{aligned}
    \label{eq5}
\end{equation}
where $\ell$ is a linear layer. After then, the features $F_t^{FocusT}$ are added by the original current frame features $\mathcal{F}_{t}$ with residual design, and then are fed into a feed-forward network. Therefore, the whole MLA block is formulated as:
\vspace{-1mm}
\begin{equation} 
    \begin{aligned}
         \hat{\mathcal{F}_{t}} &=(\mathcal{F}_t^{FocusT} + (\mathcal{F}_{t})),\\
         \mathcal{F}_{t}^{out}&={\rm FFN}(\rm LN(\hat{\mathcal{F}_{t}})) + \hat{\mathcal{F}_{t}} \\
    \end{aligned}
    \label{eq5}
\end{equation}
where $LN$ denotes a layer normalization function to increase the fitting ability of the network and FFN is a linear layer, $\mathcal{F}_t^{out}$ is the output of the current blocks. Notably, the design of our FocusTrack block preserves the computational efficiency inherent in linear attention while enabling effective modeling of important foreground targets, showcasing a commendable balance between computational efficiency and perceptual performance. Furthermore, we visualize the corresponding visualized BEV features in Fig. ~\ref{feature}, the empirical validation demonstrates the efficacy of this dynamic weighting mechanism in modeling inter-frame motion support the interpretability of our approach in 3D SOT task. 


\noindent\paragraph{\textbf{Motion Prediction Head}} We employ a multi-subtask prediction head similar to works~\cite{synctrack,p2p} to predict the target's center offset $(x, y)$, height $z$, and rotation angle $\theta$ independently. Unlike previous trackers~\cite{osp2b,zhao2024ost}, we do not apply a multi-scale feature aggregation strategy. Instead, we directly use the final features from the multi-stage backbone for these prediction tasks. The training loss is the same as~\cite{v2b}: 
\vspace{-2mm}
\begin{equation}
\mathcal{L}=\lambda_1\mathcal{L}_{(x,y)}+\lambda_2\mathcal{L}_{z}+\lambda_3\mathcal{L}_{rot}
\end{equation}
where $\lambda_1$, $\lambda_2$ and $\lambda_3$ are hyper-parameters to balance different losses. Training loss details can be referred to~\cite{v2b}.
\section{Experiments}

\begin{table*}[t]
\caption{Comparisons with state-of-the-art methods on KITTI dataset~\cite{kitti}. The upper and lower parts include two-stream and one-stream trackers, respectively. \textit{Success} / \textit{Precision} are used for evaluation. \textbf{Bold} and \underline{underline} denote the best result and the second-best one, respectively. $\dagger$ means the methods was pre-trained on large-scale images and point cloud dataset~\cite{fan2019lasot,huang2019got,pang2022masked}.}
\centering
    \resizebox{1\textwidth}{!}{
    \normalsize
    \begin{tabular}{c|cc|c|cccc|cc}
          \toprule[0.4mm]
            && & Mean& Car & Pedestrian &  Van & Cyclist &  \\
           \multirow{-2}{*}{Paradigm}&\multirow{-2}{*}{Tracker} & \multirow{-2}{*}{Source} & (14,068) & (6,424)&(6,088) & (1,248) & (308) & \multirow{-2}{*}{FPS} & \multirow{-2}{*}{Device} \\
          \midrule
          \multirow{21}{*}{Match}&SC3D~\cite{sc3d}& CVPR'19 & 31.2 / 48.5 &41.3 / 57.9   & 18.2 / 37.8 & 40.4 / 47.0 & 41.5 / 70.4   &2&GTX 1080Ti \\ 
          &P2B~\cite{p2b}& CVPR'20& 42.4 / 60.0 & 56.2 / 72.8 & 28.7 / 49.6 & 40.8 / 48.4 & 32.1 / 44.7   &40&GTX 1080Ti\\ %
        &PTT~\cite{ptt}&IROS'21 & 55.1 / 74.2 &67.8 / 81.8& 44.9 / 72.0 &43.6 / 52.5& 37.2 / 47.3 & 40&GTX 1080Ti \\ 
        &LTTR~\cite{lttr}&BMVC'21& 48.7 / 65.8 & 65.0 / 77.1 &33.2 / 56.8 & 35.8 / 45.6 &66.2 / 89.9 & 23 &GTX 1080Ti \\ 
        &MLVSNet~\cite{mlvsnet}&ICCV'21 & 45.7 / 66.6 &56.0 / 74.0   & 34.1 / 61.1 & 52.0 / 61.4 & 34.4 / 44.5  & 70 &GTX 1080Ti \\
         &BAT~\cite{bat}&ICCV'21& 51.2 / 72.8 & 60.5 / 77.7 &42.1 / 70.1 & 52.4 / 67.0 &33.7 / 45.4 & 57 &RTX 2080  \\
        &V2B~\cite{v2b}&NeurIPS'21 & 58.4 / 75.2 &70.5 / 81.3   & 48.3 / 73.5 & 50.1 / 58.0 & 40.8 / 49.7   &37&TITAN RTX \\ 
        &PTTR~\cite{pttr}& CVPR'22& 57.9 / 78.2 &65.2 / 77.4   & 50.9 / 81.6 & 52.5 / 61.8 & 65.1 / 90.5  &50 &Tesla V100 \\ 
        &STNet~\cite{stnet} & ECCV'22 & 61.3 / 80.1 &72.1 / 84.0 &49.9 / 77.2& 58.0 / 70.6& 73.5 / 93.7& 35 &TITAN RTX \\
        &CMT~\cite{cmt} & ECCV'22 &59.4 / 77.6 &70.5 / 81.9 &49.1 / 75.5& 54.1 / 64.1& 55.1 / 82.4   & 32&GTX 1080Ti \\ 
        &GLT-T~\cite{glt}&AAAI'23 & 60.1 / 79.3 &68.2 / 82.1   & 52.4 / 78.8 & 52.6 / 62.9 & 68.9 / 92.1  &30 &GTX 1080Ti \\ 
        &OSP2B~\cite{osp2b}& IJCAI'23& 60.5 / 82.3 &67.5 / 82.3   & 53.6 / 85.1 & 56.3 / 66.2 & 65.6 / 90.5   & 34& GTX 1080Ti \\ 
        &CXTrack~\cite{cxtrack} & CVPR'23& 67.5 / 85.3 & 69.1 / 81.6 &67.0 / 91.5& 60.0 / 71.8& 74.2 / 94.3& 34&RTX 3090 \\ 
        &MBPTrack~\cite{mbptrack} & ICCV'23 & 70.3 / 87.9 & \underline{73.4} / 84.8 &\underline{68.6} / \underline{93.9} &61.3 / 72.7& 76.7 / 94.3  & 50&RTX 3090  \\ 
        &SyncTrack~\cite{synctrack} &ICCV'23 & 64.1 / 81.9 &73.3 / \underline{85.0} & 54.7 / 80.5&60.3 / 70.0& 73.1 / 93.8  & 45&TITAN RTX \\ 
        &MoCUT~\cite{cutrack}&ICLR'24& 65.8 / 85.0 &67.6 / 80.5 &63.3 / 90.0 &64.5 / 78.8& \textbf{76.7} / 94.2 &48& RTX 3070Ti \\ 
        \midrule
         \multirow{4}{*}{Motion}&M$^2$Track~\cite{m2track} & CVPR'22& 62.9 / 83.4 &65.5 / 80.8   & 61.5 / 88.2 & 53.8 / 70.7 & 73.2 / 93.5  &57 &Tesla V100 \\ 
          &M$^2$Track++~\cite{m2track++} & TPAMI'23& 66.5 / 85.2 &71.1 / 82.7  & 61.8 / 88.7 & 62.8 / 78.5 & 75.9 / 94.0   &57& Tesla V100 \\ 
         &VoxelTrack~\cite{lu2024voxeltrack} & ACM MM'24& \underline{70.4} / \underline{88.3} &72.5 / 84.7 & 67.8 / 92.6  & \textbf{69.8} / \textbf{83.6} & 75.1 / \underline{94.7}   &36& TITAN RTX \\ 
         \rowcolor{myblue!18}&\textbf{FocusTrack} & \textbf{Ours}&\textbf{71.3} / \textbf{89.4} & \textbf{74.1} / \textbf{85.9}&\textbf{69.3} / \textbf{94.1} &\underline{68.4} / \underline{83.5} & \underline{75.9} / \textbf{94.7} &\textbf{105} & RTX 3090 \\
          \bottomrule[0.4mm]
    \end{tabular}
    }
\label{table1}
\vspace{-1mm}
\end{table*}

\begin{table*}[t]
\caption{Comparisons with state-of-the-art methods on nuScenes dataset~\cite{nuScenes}. \textit{Success} / \textit{Precision} are used for evaluation. \textbf{Bold} and \underline{underline} denote the best result and the second-best one, respectively.}
\centering
    \resizebox{1.0\textwidth}{!}{
    \normalsize
    \begin{tabular}{c|c|c|ccccc}
          \toprule[0.4mm]
           \multirow{2}{*}{Tracker} & Mean & Mean & Car & Pedestrian &  Truck & Trailer & Bus \\
            & (117,278) & Category & (64,159) & (33,227)&  (13,587) & (3,352) &(2,953) \\
          \midrule
          SC3D~\cite{sc3d} & 20.70  / 20.20 &25.78 / 22.90 &22.31  / 21.93& 11.29  / 12.65& 30.67  / 27.73& 35.28  / 28.12& 29.35  / 24.08\\
           P2B~\cite{p2b} & 36.48  / 45.08 & 38.41 / 40.90 & 38.81  / 43.18 &28.39  / 52.24& 42.95  / 41.59& 48.96  / 40.05& 32.95  / 27.41\\
           PTT~\cite{ptt}  & 36.33  / 41.72 & 40.38 / 41.81 &41.22  / 45.26& 19.33  / 32.03&50.23  / 48.56&51.70  / 46.50&39.40  / 36.70 \\
           BAT~\cite{bat} &38.10  / 45.71 & 40.59 / 42.42 & 40.73  / 43.29 &28.83  / 53.32& 45.34  / 42.58& 52.59  / 44.89 &35.44  / 28.01 \\
           PTTR~\cite{pttr}& 44.50  / 52.07 & 43.22 / 44.91 & 51.89  / 58.61& 29.90  / 45.09& 45.30  / 44.74 &45.87  / 38.36& 43.14  / 37.74\\
           GLT-T~\cite{glt} &44.42  / 54.33 & 47.03 / 50.98 &48.52  / 54.29&31.74  / 56.49& 52.74  / 51.43&57.60  / 52.01& 44.55  / 40.69\\
           MoCUT~\cite{cutrack} & 51.19 / 64.63 & 54.17 / 62.28 &57.32 / 66.01 &33.47 / 63.12& 61.75 / 64.38 &60.90 / 61.84 &57.39 / 56.07\\
           MBPTrack~\cite{mbptrack} & 57.48  /  69.88 & 58.10 / 64.19  & 62.47  /  70.41 & 45.32  /  {74.03} &  62.18  /  63.31 &  65.14  /  61.33&   55.41  /  51.76 \\
           \midrule
           M$^2$Track~\cite{m2track} & 49.23  / 62.73 & 50.86 / 59.05  &55.85  / 65.09& 32.10  / 60.92 &57.36  / 59.54& 57.61  / 58.26& 51.39  / 51.44\\ 
           P2P-voxel~\cite{p2p}& \underline{59.84} / \underline{72.13} & \underline{61.04} / \underline{67.44} & \underline{65.15} / \underline{72.90} & \underline{46.43} / \underline{75.08} & \underline{64.96} / \underline{65.96} & \underline{70.46} / \underline{66.86} & \underline{59.02} / \underline{56.56} \\
            \rowcolor{myblue!18}\textbf{FocusTrack (Ours)} &  \textbf{61.72}  /  \textbf{74.35} & \textbf{64.50} / \textbf{70.95} &   \textbf{66.50}  /  \textbf{74.37}& \textbf{47.85}  /  \textbf{77.53}& \textbf{69.34}  /  \textbf{69.98}& \textbf{74.76}  /  \textbf{71.20}& \textbf{64.02}  /  \textbf{61.67}\\
          \bottomrule[0.4mm]
  \end{tabular}}
\label{table3}
\vspace{-2mm}
\end{table*}

\subsection{Experiment Setting}

\noindent\paragraph{\textbf{Implementation Details}}
The extended ranges of input point cloud regions are defined as extended range [(-4.8,4.8),(-4.8,4.8),(-1.5,1.5)] and [(-1.92,1.92),(-1.92,1.92),(-1.5,1.5)] to contain relevant points for cars and humans, respectively. We first crop out point cloud regions from consecutive point clouds $\mathcal{P}_{t-1}$ and $\mathcal{P}_t$ that have the same aspect ratio as the target and are 2 times the target box size. Then, a resolution-consistent pillarization operation with spatial resolution of $128\times 128$ is applied to structurally normalize the cropped point cloud. The spatial resolution after pillar feature encoder is $128\times128\times1$. After the FocusTrack backbone, the spatial resolution of BEV features $F_{t-1}^{bev}$ are downsampling to $16\times 16\times 128$. 

\noindent\paragraph{\textbf{Datasets and Evaluation Metrics}}
To evaluate our proposed FocusTrack, we conduct extensive experiments on KITTI~\cite{kitti}, nuScene-s~\cite{nuScenes} and Waymo Open Dataset (WOD)~\cite{waymo}. To be fair, we follow common setup~\cite{p2b,ptt,bat,m2track, pttr,gltt,mbptrack,cutrack} of training and testing models on the KITTI and nuScenes datasets. Besides, we also follow previous methods~\cite{v2b,stnet,cxtrack,cutrack,pre-train3D} and apply pre-trained KITTI model to the nuScene and WOD dataset for testing to evaluate the generalization ability. Note that, nuScenes and WOD are more challenging than KITTI due to large data volume and complex scenes. In addition, KITTI and WOD data are collected by 64-beam LiDAR sensors, while nuScenes data are captured by 32-beam LiDARs. Therefore, point cloud data of nuScenes are more sparse than KITTI and WOD. For evaluation metrics, we employ one pass evaluation (OPE)~\cite{otb2013,kristan2016novel} to evaluate tracking performance in terms of Success and Precision, which is the common setup~\cite{ptt,bat,p2b,m2track,pttr,glt,mbptrack}. Please refer to appendix for more implementation details.

\begin{table*}[!htbp]
\caption{Generalization comparisons on Waymo Open Dataset (WOD)~\cite{waymo}. All the model is pre-trained on KITTI and evaluated on WOD \textit{val} set. \textit{Success} / \textit{Precision} are used for evaluation. \textbf{Bold} and \underline{underline} denote the best result and the second-best one.}
\vspace{-2mm}
\centering
    \resizebox{1.0\textwidth}{!}{
    \begin{tabular}{c|c|cccc|cccc}
          \toprule[0.4mm]
         & &\multicolumn{4}{c}{Vehicle} & \multicolumn{4}{c}{Pedestrian}\\
         & & Easy& Medium & Hard & Mean & Easy & Medium & Hard & Mean \\
        \multirow{-3}{*}{Tracker} & \multirow{-3}{*}{Mean} & (67,832)&(61,252)&(56,647)&(185,731)& (85,280)&(82,253)&(74,219)&(241,752) \\
          \midrule[0.4mm]
          P2B~\cite{p2b}& 33.0 / 43.8 & 57.1 / 65.4 & 52.0 / 60.7 & 47.9 / 58.5 & 52.6 / 61.7 & 18.1 / 30.8 & 17.8 / 30.0 & 17.7 / 29.3 & 17.9 / 30.1 \\ 
          BAT~\cite{bat}& 34.1 / 44.4 & 61.0 / 68.3 & 53.3 / 60.9 & 48.9 / 57.8 & 54.7 / 62.7 & 19.3 / 32.6 & 17.8 / 29.8 & 17.2 / 28.3 & 18.2 / 30.3 \\
          V2B~\cite{v2b}& 38.4 / 50.1 & 64.5 / 71.5 & 55.1 / 63.2 & 52.0 / 62.0 & 57.6 / 65.9 & 27.9 / 43.9 & 22.5 / 36.2 & 20.1 / 33.1 & 23.7 / 37.9 \\
         STNet~\cite{stnet}&40.4 / 52.1 & {65.9} / {72.7}& {57.5} / {66.0}& {54.6} / {64.7} &{59.7} / {68.0}& 29.2 / 45.3& 24.7 / 38.2& 22.2 / 35.8 &25.5 / 39.9  \\
    CXTrack~\cite{cxtrack} & 42.2 / 56.7 & 63.9 / 71.1 &54.2 / 62.7& 52.1 / 63.7 &57.1 / 66.1& 35.4 / 55.3 &29.7 / 47.9& 26.3 / 44.4 &30.7 / 49.4 \\
    MoCUT~\cite{cutrack}&45.2 / 58.5 &68.3 / 75.0 &59.4 / 66.9 &57.1 / 66.3 &61.9 / 69.7 &36.5 / 54.8 &30.8 / 48.9 &29.5 / 45.4 &32.4 / 49.9\\
    MBPTrack~\cite{mbptrack} &46.0 / 61.0 & \underline{68.5} / \underline{77.1} &\underline{58.4} / \underline{68.1}& \underline{57.6} / \underline{69.7}& \underline{61.9} / \underline{71.9} &{37.5} / {57.0} &{33.0} /{51.9}& {30.0} / {48.8}& {33.7} /{52.7} \\
    \midrule[0.4mm]
    M$^2$Track~\cite{m2track}&44.6 / 58.2 &68.1 / 75.3& 58.6 / 66.6 &55.4 / 64.9 &61.1 / 69.3& 35.5 / 54.2 &30.7 / 48.4 &29.3 / 45.9& 32.0 / 49.7 \\
    P2P-voxel~\cite{p2p} &\underline{47.2} / \underline{62.9} & 66.2 / 73.8 & 57.8 / 67.0 & 56.8 / 68.1 & 60.0 / 69.1 &\underline{43.7} / \underline{65.2} &\underline{36.4} / \underline{57.1}& \underline{31.3} / \underline{51.0}& \underline{37.4} /\underline{58.1} \\
    \rowcolor{myblue!18} \textbf{FocusTrack (Ours)} &\textbf{49.5} / \textbf{65.6} &  \textbf{69.2} / \textbf{77.9}&\textbf{59.5} / \textbf{69.3}&\textbf{58.7} / \textbf{70.8} &\textbf{62.8} / \textbf{72.9} & \textbf{44.5} / \textbf{66.1} &\textbf{38.5} / \textbf{59.4}&\textbf{33.9} / \textbf{53.5} &\textbf{39.3} / \textbf{60.0} \\ 
    \bottomrule[0.4mm]
    \end{tabular}}
\vspace{-3mm}
\label{table2}
\end{table*}

\vspace{-1mm}
\subsection{Comparison with State-of-the-art Trackers}
\vspace{-3mm}
\noindent\paragraph{\textbf{Results on KITTI}} We present comprehensive comparisons of tracking performance on KITTI~\cite{kitti} dataset, as well as computation efficiency. As shown in Tab.~\ref{table1}, FocusTrack exhibits superior performance across various categories, achieving the highest mean Success and Precision rates of 71.3\% and 89.4\%, respectively. Moreover, FocusTrack outperforms the previous leading match based method, i.e., MBPTrack~\cite{mbptrack} by 1.0\% and 1.5\%, while demonstrating significant advantages in terms of computation efficiency, achieving 2.24X faster running speed of 105 FPS. Compared to the top-performing motion tracker M$^2$Track++~\cite{m2track++}, FocusTrack achieves remarkable performance improvements in all categories, surpassing it by up to 4.8\% and 4.2\%. This proves the effectiveness of our FocusTrack framework, which effectively explores motion cues for tracking.

\vspace{-1.3mm}
\noindent\paragraph{\textbf{Results on nuScenes}} We also conduct experiments on the large-scale nuScenes~\cite{nuScenes} dataset. NuScenes contains many complex scenes with sparser point clouds, making it a more challenging benchmark. As shown in Tab.~\ref{table3}, our FocusTrack outperforms all previous match and motion based trackers across all categories. Notably, FocusTrack exhibits considerable leading performance, surpassing the state-of-the-art tracker MBPTrack by 4.24\% and 4.47\%, respectively. This demonstrates the inherent advantages our proposed FocusTrack framework, which has the capacity to deliver exceptional model performance on large-scale datasets. Furthermore, given the complex scenes and more diverse tracking categories in nuScenes, the superior performance also shows the tremendous potential of our framework for practical applications.

\vspace{-2mm}
\subsection{Generalization Experiments}
\label{generalize}
\vspace{-2mm}

\noindent\paragraph{\textbf{Generalization Results on Waymo}} 
To validate the generalization ability of the proposed method, we follow the previous methods~\cite{stnet,cxtrack,pre-train3D,v2b,mbptrack,p2p} setting and conduct an evaluation by applying the Car and Pedestrian models trained on KITTI dataset to the WOD~\cite{waymo} dataset, following comment setting~\cite{v2b,mbptrack}. As shown in Tab.~\ref{table2}, our FocusTrack outperforms the other comparison methods across all categories, particularly in the Pedestrian category where it surpasses SOTA dual-stream tracker MBPTrack by up to 5.6\% and 7.3\%. This indicates the strong generalization of the proposed framework to unseen scenes.

\subsection{Ablation Study}
\vspace{-2mm}

\noindent\paragraph{\textbf{Overall Ablation Comparison}} We conduct an ablation study to investigate the effect of the series of CNN, linear layers and DWC and the proposed IMM module. As shown in Tab.~\ref{tab:kitti_ablation}, by incorporating all macro architecture design, the model achieves performance improvements across different setting. Specifically, the DWC ((b) to (c)) lead to 1.4\%/1.7\% and 1.7\%/1.7\% improvement in car and pedestrian categories, respectively. Furthermore, when using the IMM module, the model performs substantial accuracy gains of 2.5\%/3.8\% and 2.9\%/3.1\%. The results demonstrate that our IMM module effectively captures inter-frame motion differences, facilitating for the subsequent attention to model motion offset.

\begin{table}[!htbp]
\caption{Ablation results of different module on the KITTI \textit{val} set for car and pedestrian.}
\centering
\resizebox{1.0\linewidth}{!}{%
\normalsize
\begin{tabular}{c|c|c|c|c|cc}
\hline\noalign{\smallskip}
Setting &CNN & Linear & DWC & IMM & Car & Pedestrian \\
\hline
\noalign{\smallskip}
(a)&\cc & \xx & \xx &\xx &69.4 / 79.6 & 63.1 / 87.7\\
(b)&\cc & \cc & \xx& \xx&70.2 / 80.4&64.5 / 89.3 \\
(c)&\cc & \cc &  \cc&\xx &71.6 / 82.1&66.2 / 91.0\\
\rowcolor{myblue!18} (d)&\cc & \cc &  \cc  &  \cc & \textbf{74.1} / \textbf{85.9}&\textbf{69.3} / \textbf{94.1}\\
\hline
\end{tabular}
}
\label{tab:kitti_ablation}
\vspace{-2mm}
\end{table}

\begin{figure}[!htbp]
\centering
\includegraphics[width=1.0\linewidth]{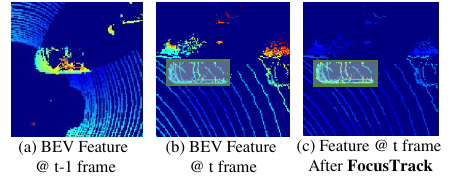}
\vspace{-4mm}
\caption{The BEV feature visualization of FocusTrack.}
\label{feature}
\vspace{-3mm}
\end{figure}

\vspace{-2mm}
\noindent\paragraph{\textbf{Feature Visualization of FocusTrack}} Fig.~\ref{feature} shows that the visualizations of the attention feature map with proposed FocusTrack. From Fig.~\ref{feature}~(b) to (c), the background noise in the current frame is effectively suppressed and foreground points are gained more attention. This visualization analysis of the feature map provides salient evidence supporting the key contributions of the FocusTrack.

\begin{table}[h]
\vspace{-2mm}
\caption{Ablation of w/o shared setting for CNN and Linear layers in FocusTrack block on the KITTI dataset.}
\vspace{-2mm}
\centering
\resizebox{1.0\linewidth}{!}{
\normalsize
\begin{tabular}{c|c|cc|c}
\hline\noalign{\smallskip}
Setting & Shared & Car & Pedestrian & FPS \\
\noalign{\smallskip}
\hline
(a) &\xx & 73.2 / 83.8 & 68.7 / 92.9 & 91\\
\rowcolor{myblue!18} (b) & \cc &\textbf{74.1} / \textbf{85.9}&\textbf{69.3} / \textbf{94.1} & \textbf{105}\\
\hline
\end{tabular}
}
\label{tab:kitti_shared}
\vspace{-4mm}
\end{table}

\noindent\paragraph{\textbf{Effectiveness of Weight-Shared Design}} We conduct an ablation study to investigate whether the series of CNN, linear layers and DWC designs in the IMM module should be weight-shared. As shown in Tab.~\ref{tab:kitti_shared}, the shared design results in improved performance and faster speed. Compared to the non-shared design (a), it achieves performance gains of 0.9\%/1.9\% and 0.6\%/1.2\% in car and pedestrian classes, while speed increases from 91 to 105 FPS. We think that this Siamese-like sharing mechanism can better extract salient features from inputs at different time steps under the same weights, benefiting inter-frame motion difference modeling.

\begin{table}[!htbp]
\vspace{-3mm}
\caption{Ablation study on downsample ratio.}
\vspace{-2mm}
\label{stage_number}
\centering
    \resizebox{1.0\linewidth}{!}{
    \normalsize
    \begin{tabular}{c|c|cc|c}
          \toprule[0.4mm]
          Ratio&Resolution& Car & Pedestrian &FPS\\
          \midrule
           2X &  64 X 64 &71.5 / 82.1 &  66.1 / 90.1  &\textbf{118} \\
           4X & 32 X 32&  72.4 / 81.9 &  67.2/ 91.5  & 112 \\
           \rowcolor{myblue!18} 8X & 16 X 16&\textbf{74.1} / \textbf{85.9}&\textbf{69.3} / \textbf{94.1} &105 \\
           16X & 8 X 8&73.6 / 83.6 & 68.4 / 92.5 & 98\\
          \bottomrule[0.4mm]
    \end{tabular}}
    \vspace{-4mm}
\end{table}

\noindent\paragraph{\textbf{The Downsample Ratio in Backbone.}} The backbone needs extract different level discriminative geometrics. Our FocusTrack backbone boosts high-level semantics by adopting downsampling at each stage. Naturally, the stage number also affects the performance. Here, we compare the tracking accuracy when feature extraction is performed under different downsampling ratios. As shown in Tab.~\ref{stage_number}, the 8× down-sampling configuration yields the best performance. 

\noindent\paragraph{\textbf{The Learning Parameter $\alpha$.}} In the Inter-frame Motion Modeling (IMM) module, the learnable parameter $\alpha$ controls the magnitude of inter-frame motion differences. When $\alpha$ approaches 0, the previous frame has minimal influence on the current frame; when $\alpha$ equals 1, the model simplifies to a basic inter-frame difference. We analyze $\alpha$ variations across backbone stages for cars and pedestrians in the nuScenes dataset, as shown in the Figure. For cars, $\alpha$ values are 0.21, 0.39, and 0.53; for pedestrian, they are 0.24, 0.43, and 0.61. As the network deepens, higher $\alpha$ values indicate greater emphasis
\begin{wrapfigure}{r}{4.3cm}
  \vspace{-2mm}
    \includegraphics[width=1.0\linewidth]{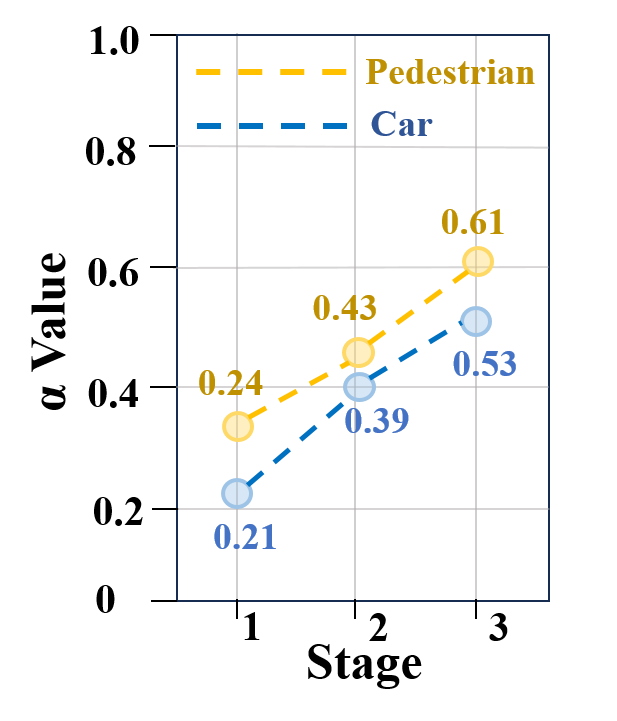}
  \vspace{-8mm}
\end{wrapfigure}
on dynamic motion patterns. Pedestrians consistently show higher $\alpha$ values, reflecting their more variable motion, while cars exhibit lower values, indicating more stable motion. This variation highlights the model's capacity to flexibly adjust its attention to motion cues based on the distinct behaviors of different object categories, thereby enhancing tracking performance in diverse scenarios.

\begin{figure}[t]
\centering
\includegraphics[width=1.0\linewidth]{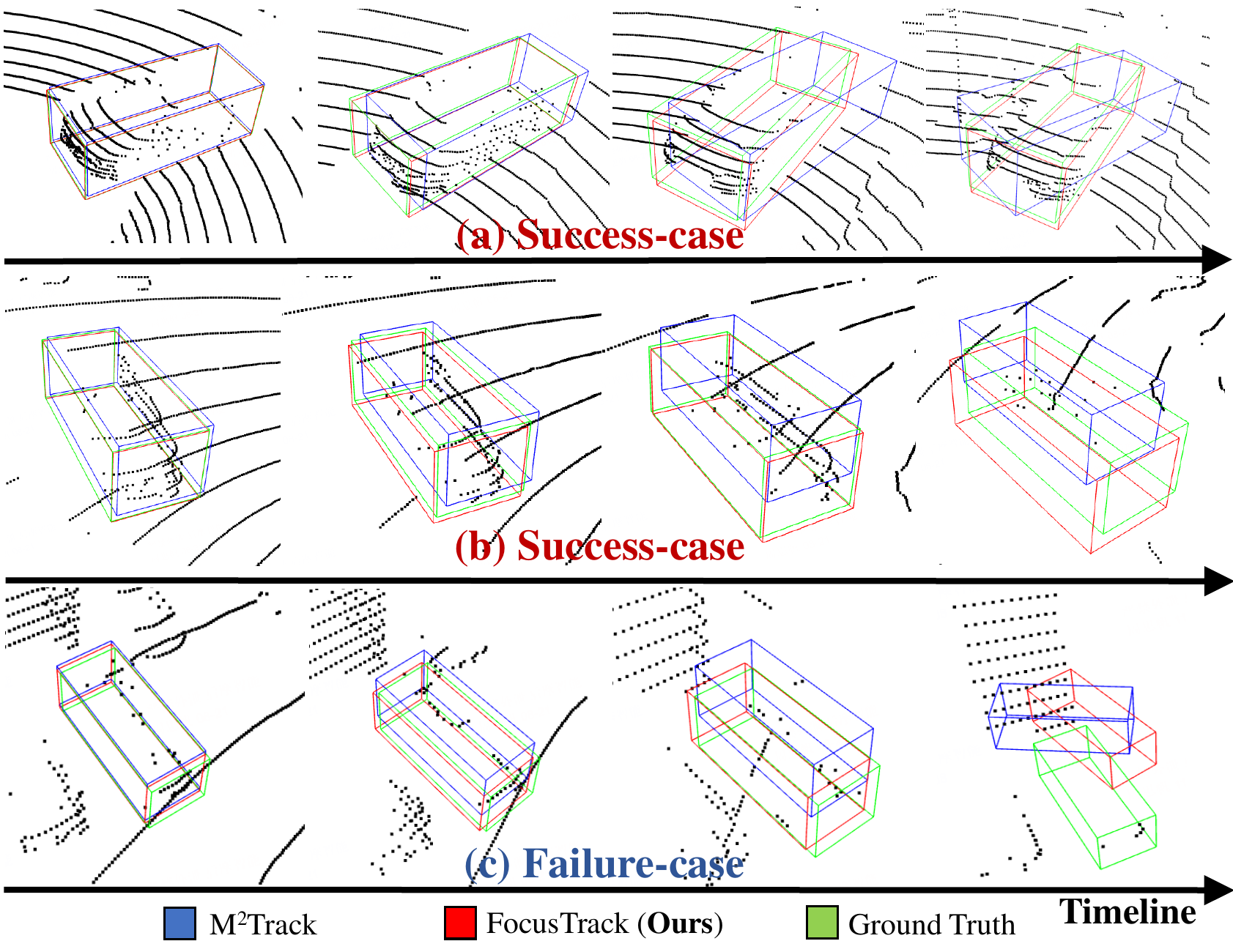}
\vspace{-4mm}
\caption{Visualization of tracking results compared with state-of-the-art motion-based M2Track~\cite{m2track} method.}
\label{vis_fig}
\vspace{-4mm}
\end{figure}

\subsection{Visualization Results}
\vspace{-2mm}
\noindent\paragraph{\textbf{Success Case}} As shown in Fig.~\ref{vis_fig}, we visualize tracking results over the SOTA method M2Track~\cite{m2track} on nuScenes~\cite{nuScenes} dataset, across diverse trajectories. Whether in dense scenarios (Fig.~\ref{vis_fig} (a)), or sparse scenarios (Fig.~\ref{vis_fig} (b)), our FocusTrack is able to track the target, while M2Track performs inaccurate bounding box estimation. This highlights the superior tracking performance of our FocusTrack across a range of conditions, from dense to sparse scenes.

\vspace{-1mm}
\noindent\paragraph{\textbf{Failure Case}} While our method demonstrates effective performance, as shown in Fig.~\ref{vis_fig} (c), our method is inherently presents limitations in sparse scenes. The low resolution of LiDAR point cloud and its focus on capturing geometry-only information restrict performance in such extreme-sparse scenarios. Naturally, this challenge could be effectively addressed by integrating dense RGB images, which would provide richer textual details. We consider this an open problem for the future work.
\section{Conclusion}
\vspace{+1mm}
In this work, we propose a novel one-stage paradigms tracking framework that unifies motion-semantics co-modeling. The core components of our DiffTrack are the Inter-frame Motion Modeling (IMM) module and the Focus-and-Suppress Attention. The former module capture the global motion differences in successive frames. The later module aims to extract intra-frame spatial geometrics while enhancing the foreground semantics based on the temporal-aware motion context from IMM module. Based on above two designs, FocusTrack enables end-to-end training with
compact one-stage pipelin. Extensive experiments on KITTI~\cite{kitti}, nuScenes~\cite{nuScenes} and WOD~\cite{waymo} datasets demonstrate that our FocusTrack sets a new state-of-the-art performance, while running at a high speed of 105 FPS on a single RTX3090 GPU. Finally, our FocusTrack provides a better trade-off between efficiency and accuracy in latency-sensitive tracking tasks based on our "focus-and-suppress" design principle. 

\section{Limitation and Future Work}

Our FocusTrack framework currently faces a limitation in sparse scenarios, where performance is suboptimal—a challenge that is commonly observed in existing methods~\cite{m2track,m2track++,p2p}. An effective solution would be to integrate temporal information by leveraging historical global trajectories over a past duration alongside local differential motion data or fuse dense RGB images~\cite{MVCTrack} to enhance the dense semantic representation. In future work, we plan to incorporate multi-frame temporal information to capture the dynamic motion of tracked objects. We will also explore richer structured cues for sparse-scene robustness~\cite{liu2026health} and explainable motion modeling for better transparency~\cite{shen2025aienhanced}. Besides, we will also explore the quantization-aware method~\cite{zhoulidarptq,gsq,jiang2025ptq4ris,wang2025point4bit,xumambaquant,hu2025ostquant,yu2025mquant,chenmoequant,yuerwkvquant,yu2025fq} to accelerate 3D Tracker for onboard applications and the integration with privacy-preserving federated learning~\cite{liu2024fedbcgd,liu2025consistency,liu2025improving}.

\section*{Acknowledgement}
We thank all anonymous reviewers, ACM MM Program Committee, and Area Committee for their kind help of this work. This work was supported by the National Natural Science Foundation of China (No.62271143), Frontier Technologies R\&D Program of Jiangsu (No. BF2024060) and the Big Data Computing Center of Southeast University.
\bibliographystyle{ACM-Reference-Format}
\bibliography{sample-base}
\appendix
\section{More Generalization Experiments}
\subsection{Generalization ability on Different Classes.}
\label{seca.3}
To further validate the the generalization of the proposed class-agnostic model, we conduct a series of experiments using KITTI~\cite{kitti} and nuScenes~\cite{nuScenes} datasets.

\begin{table}[h]
\centering
    \resizebox{0.9\linewidth}{!}{
    \normalsize
    \begin{tabular}{c|cc}
          \toprule[0.4mm]
           Seen Classes $\to$ Unseen Classes & Van &Cyclist \\
          \midrule
          Van, Cyclist (Tab. 1) & 68.4 / 83.5 & 75.9 / 94.7\\
         Car,Pedestrian$\to$Van,Cyclist & 65.8 / 79.4 &73.1 / 91.8 \\
          \bottomrule[0.4mm]
    \end{tabular}}
\caption{Generalization experiments from seen object classes to unseen ones on KITTI~\cite{kitti} dataset.}
\label{tab9}
\end{table}
\vspace{-2mm}
First, we conduct a generalization experiment by training our FocusTrack on certain classes of data and testing it on unseen classes of data. As shown in Tab.~\ref{tab9}, FocusTrack is trained on Car and Pedestrian classes, which is applied for Van and Cyclist classes for test. Compared to performance in normal setup in Tab. 1 in the main paper, only a slight degradation in performance is observed. The results demonstrate that our FocusTrack is able to generalize to data of unseen classes.

\begin{table}[h]
\centering
    \resizebox{0.9\linewidth}{!}{
    \normalsize
    \begin{tabular}{c|cc}
          \toprule[0.4mm]
           nuScenes$\to$KITTI nuScenes & Car &Pedestrian\\
            \midrule
            KITTI (Tab. 1) & 74.1 / 85.9 & 69.3 / 94.1\\
            nuScenes$\to$KITTI & 58.6 / 72.3 &63.1 / 88.9 \\
          \bottomrule[0.4mm]
    \end{tabular}}
\caption{Generalization experiments between KITTI~\cite{kitti} and nuScenes~\cite{nuScenes} datasets with distinct point cloud sparsity.}
\label{tab10}
\end{table}
\vspace{-2mm}
\subsection{Generalization ability on Different Datasets}
Second, we test the trained model on another dataset. Tab.~\ref{tab10} presents generalization performance under such a setup. Since nuScenes data are more sparse than KITTI data, it is challenging to generalize across the two datasets. Nevertheless, our FocusTrack generalizes well from nuScenes to KITTI, \textit{i.e.}, sparse point clouds to dense point clouds, obtaining 72.3\% and 88.9\% mean Precision in terms of Car and Pedestrian classes. 

\section{More Implementation Details}
\subsection{Dataset and Evaluation Metrics} 
We evaluate our method on KITTI~\cite{kitti}, nuScenes~\cite{nuScenes} and Waymo Open Dataset (WOD)~\cite{waymo}.  To be fair, we follow common setup~\cite{p2b,m2track,mbptrack} of training and testing models on the KITTI and nuScenes datasets, and applying pre-trained KITTI model to the WOD dataset for testing. Note that, nuScenes and WOD are more challenging than KITTI due to large data volume and complex scenes. In addition, KITTI and WOD data are collected by 64-beam LiDAR sensors, while nuScenes data are captured by 32-beam LiDARs. Therefore, point cloud data of nuScenes are more sparse than KITTI and WOD.

\noindent\textbf{KITTI.} KITTI~\cite{kitti} contains 21 training sequences and 29 test sequences. Due to test labels are not available, we split the training sequences into train [0-17), val [17-19) and test [19, 21) sets following common setting~\cite{sc3d,p2b,bat,pttr,m2track,cxtrack,mbptrack,cutrack,pttr++}.

\noindent\textbf{NuScenes.} NuScenes~\cite{nuScenes} involves 1000 scenes, which are divided into 700, 150 and 150 scenes for train, val and test, respectively. Following~\cite{bat,m2track,mbptrack}, we train our FocusTrack using ``train$\_$track'' split and test it on val set. 

\noindent\textbf{Waymo.} Waymo Open Dataset (WOD)~\cite{waymo} comprises 1121 tracklets categorized into easy, medium and hard sub-sets based on the point cloud sparsity. We follow LiDAR-SOT~\cite{lidarsot,v2b,stnet} to evaluate our model. 

\noindent\textbf{Evaluation Metrics.}
Following common practice, we employ One Pass Evaluation (OPE) ~\cite{otb2013,kristan2016novel} to evaluate tracking performance using both Success and Precision metrics. Success calculates the intersection over union (IOU) between the predicted bounding box and the ground truth one, while Precision assesses the distance between the centers of the two corresponding bounding boxes.

\vspace{-3mm}
\subsection{Model Details}
We first crop out point cloud regions from consecutive point clouds $\mathcal{P}_{t-1}$ and $\mathcal{P}_t$ that have the same aspect ratio as the target and are 2 times the target box size. Then, we apply a pillar feature encoding to the cropped point cloud regions to obtain BEV features $\mathcal{F}_{t-1}$ and $\mathcal{F}_{t}$ with a spatial resolution of $128 \times 128 \times 16$, in order to structurally normalize the point cloud regions. The spatial resolution after pillar feature encoder is $128\times128$. These tokens are then fed into 3 sequential FocusTrack blocks, where each FocusTrack block downsample the spatial resolution while doubling the number of channels, to focus on the foreground tracking objects based on the differential motion cues. After the FocusTrack backbone, the spatial resolution are downsampled to $16\times 16\times 128$. In the prediction head, we use 3 convolution blocks to spatially downsample the output features and increase the number of channels to obtain features with a dimension of $1 \times 1 \times 512$. These features are then flattened and input a MLP to get the final output. 

\vspace{-3mm}
\subsection{Data Augmentation.}
During training stage, we introduce simulated test errors for consecutive frames to enhance model's accuracy. In contrast to previous methods~\cite{p2b,m2track,mbptrack,cutrack} that crop point cloud regions in $t-1$ and $t$ frame centered on target boxes in corresponding frame, respectively, we use target box in $t-1$ frame as the center of the cropped point cloud regions in both $t-1$ and $t$ frame. By this way, the model is able to learn tracking errors occurred in testing stage. In addition, we randomly flip target points and box horizontally and rotate it uniformly around its longitudinal axes in the range [-5$^\circ$,5$^\circ$].

\subsection{Training \& Inference}
\noindent\textbf{Training.} 
We train our tracking models using the AdamW optimizer on a NVIDIA RTX 4090 GPU, with a batch size of 128. The initial learning rate is set to 1e-4 and is decayed by a factor of 5 every 20 epochs.

\begin{algorithm}[H]
\caption{Inference process of FocusTrack in a point cloud sequence}
\label{al1}
\textbf{Input:} $T$-frame sequence $\{\mathcal{P}_t\}_{t=1}^{T}$; Tracked target $\mathcal{B}=(x,y,z,w,h,l,\theta)$ in the 1st frame. \\
\textbf{Output:} Tracking results \centering $\{\mathcal{B}_t=(x_t,y_t,z_t,\theta_t)\}_{t=2}^{T}$

\begin{algorithmic}[1]
\FOR{$t = 2$ to $T$}
    \STATE Crop point cloud regions $\mathcal{P}_{t-1}^{crop}$ and $\mathcal{P}_{t}^{crop}$.
    \STATE Pillarize regions in a spatial resolution of $\mathbb{R}^{128 \times 128 \times 16}$.
    \STATE Extract features and interact between $F_{t-1}$ and $F_{t}$ using the FocusTrack block to obtain the final $F_{t} \in \mathbb{R}^{16 \times 16 \times 128}$.
    \STATE Predict relative motion $[\Delta x, \Delta y, \Delta z, \Delta \theta]_{t-1 \to t}$ based on the final $F_{t}$ using the prediction head.
    \STATE Based on $[\Delta x, \Delta y, \Delta z, \Delta \theta]_{t-1 \to t}$ to transform the previous $\mathcal{B}_{t-1}$ and output $\mathcal{B}_t$.
\ENDFOR
\STATE \textbf{Return} $\{\mathcal{B}_t\}_{t=2}^{T}$.
\end{algorithmic}
\end{algorithm}

\noindent\textbf{Inference.}
During inference stage, FocusTrack continuously regresses out target relative motion $[\Delta x,\Delta y,\Delta z,\Delta\theta]_{t-1\to t}$ frame-by-frame. The regressed relative motion is then applied to target box $\mathcal{B}_{t-1}$ in previous frame to locate the target box $\mathcal{B}_{t}$ in current frame. Alg.~\ref{al1} presents the whole inference process of FocusTrack in a point cloud sequence.


\end{document}